\RequirePackage{fix-cm}
\documentclass[twocolumn]{svjour3}          

\smartqed  
\usepackage{graphicx}
\usepackage{enumitem}
\usepackage [autostyle, english = american]{csquotes}
\MakeOuterQuote{"}
\usepackage{hyperref}
\hypersetup{
    colorlinks=true,
    linkcolor=blue,
    filecolor=magenta,      
    urlcolor=cyan,
    citecolor=blue
}

\usepackage{amsmath}  
\usepackage{amssymb}
\usepackage{multirow}
\usepackage{booktabs}
\usepackage{subfigure}
\usepackage{epstopdf}
\usepackage{textcomp}
\usepackage{enumitem}
\usepackage{color,soul} 
\usepackage{pifont}
\usepackage[makeroom]{cancel}
\usepackage[round]{natbib}
 \setcitestyle{aysep={}} 
\usepackage{csquotes}
\usepackage[utf8]{inputenc}
\usepackage{marginnote} 

\begin{document}
\title{Attention-based Active Visual Search for Mobile Robots}

\author{Amir Rasouli \and Pablo Lanillos \and Gordon Cheng \and John K. Tsotsos
}

\institute{Amir Rasouli and John K. Tsotsos\at
              The Department of Electrical Engineering and Computer Science and Center for Vision Research, York University, Toronto, Canada \\
              Tel.: +1-416-736-2100\\
              Fax: +1-416-736-5872\\
              \email{\{aras,tsotsos\}@eecs.yorku.ca} 
           \and
           Pablo Lanillos and Gordon Cheng  \at
             The Institute for Cognitive Systems (ICS),  Technische Universit\"at M\"unchen, Institute for Cognitive Systems, Arcisstraße 21 80333 M\"unchen, Germany\\
             Tel.:+49-89-289-26818\\
             Fax: +49-89-289-26815\\
             \email{\{p.lanillos,gordon\}@tum.de}\\
              {This work has been submitted to Autonomous Robots.}
}

\date{Received: date / Accepted: date}

\maketitle

\begin{abstract}
We present an active visual search model for finding objects in unknown environments. The proposed algorithm guides the robot towards the sought object using the relevant stimuli provided by the visual sensors. Existing search strategies are either purely reactive or use simplified sensor models that do not exploit all the visual information available. In this paper, we propose a new model that actively extracts visual information via visual attention techniques and, in conjunction with a non-myopic decision-making algorithm, leads the robot to search more relevant areas of the environment. The attention module couples both top-down and bottom-up attention models enabling the robot to search regions with higher importance first. 

The proposed algorithm is evaluated on a mobile robot platform in a 3D simulated environment. The results indicate that the use of visual attention significantly improves search, but the degree of improvement depends on the nature of the task and the complexity of the environment. In our experiments, we found that performance enhancements of up to 42\% in structured  and  38\% in highly unstructured cluttered environments can be achieved using visual attention mechanisms.
\end{abstract}

\begin{figure}[!t]
\centering
\includegraphics[width=\columnwidth]{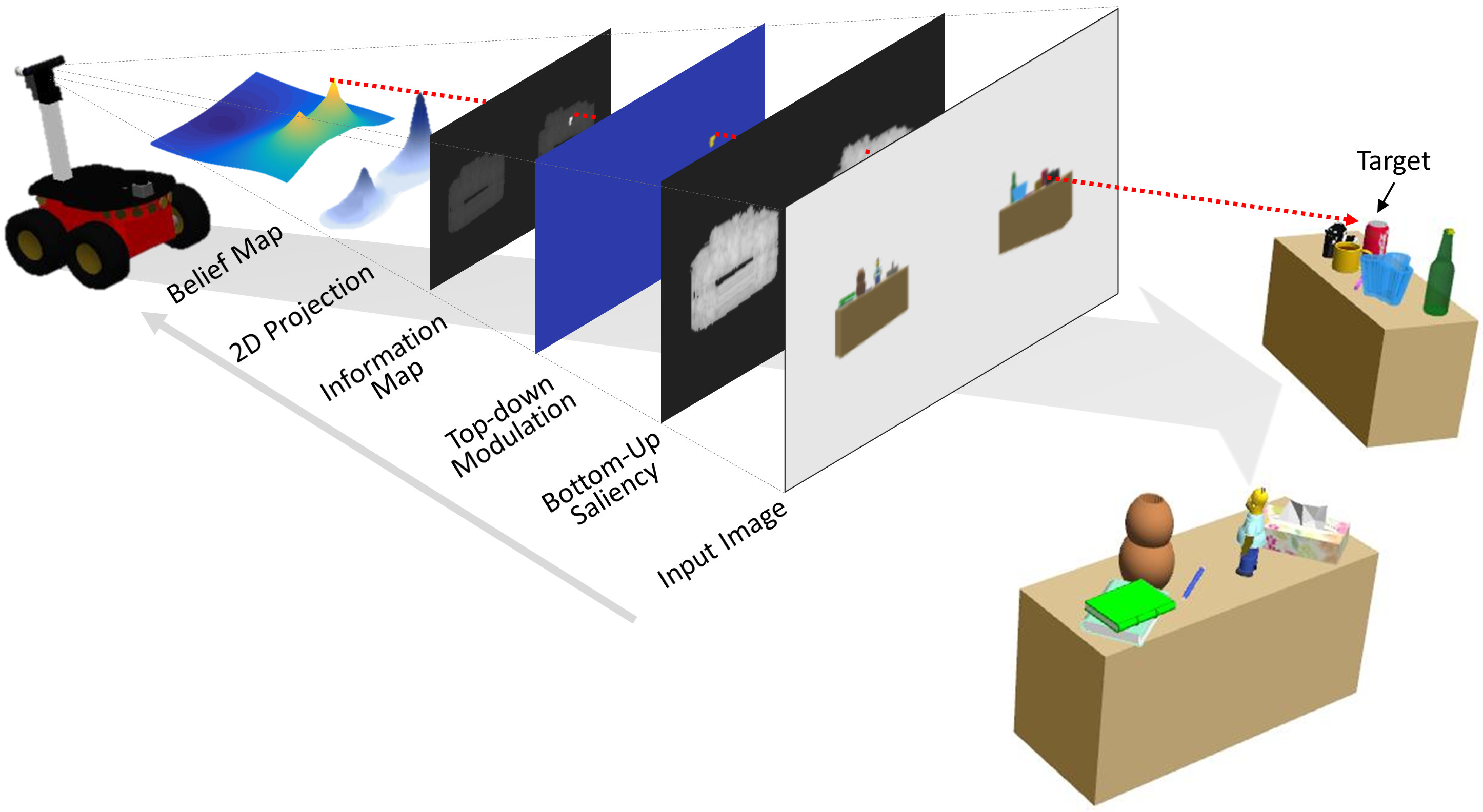} 
\caption{Attention-based active visual search. The robot, driven by the relevant visual stimuli, autonomously searches for the objects in the scene. The relevance of this visual information is defined by an attention system that combines two processes: stimulus-based (bottom-up) and goal-directed (top-down) knowledge about the object (e.g. color histogram).  In case when the sought object is the "red" can, the model ensures that the robot will first select the actions to search the surface of the table on its left.}
\label{fig:abstract}
\end{figure}

\section{Introduction}
\label{sec:intro}
Visual search is a vital ability in animals for finding food and avoiding predators, and in humans it is used in everyday life and for tasks such as natural disaster monitoring, inspections or medical image representation   \citep{eckstein2011visual,tsotsos1990analyzing}. Unfortunately, machines do not yet achieve a level of performance that matches the ability of humans in the majority of these visual search tasks, due to the difficulty of replicating the cognitive processes involved \citep{tsotsos1990analyzing,wolfe2007guided}. This, in particular, is true in applications such as finding an object in unknown environments \citep{lanillos2013mts}. Here, although a brute-force approach can solve the problem, without the use of attentive processes, the process can be very time consuming and inefficient. Attention plays an important role in managing the vast amount of information that is provided by the sensors. Therefore, it is important to incorporate, in a meaningful way, an attention model into the machine visual system to make informed decisions during an active overt visual search.

According to Stone \citep{Stone1975}, omitting the useful information provided by the visual sensors is similar to searching for an object with eyes closed. Although the object will eventually be found, the process can be very slow. In the context of visual search, visual stimuli are a valuable source of information regarding the object's whereabouts and should be actively used during the visual search. An illustrative example of searching for an object using its color property is as follows: if the robot is looking for a red can (Fig. \ref{fig:abstract}), it would be unproductive to search in places where the objects are blue.

In non-visual search, there are algorithms that couple  sensory stimuli with  control actions to reach a pursued location. For instance, for locating continuous odor sources, gradient driven techniques (e.g. chemotaxis) can be applied to compute the next actions. As the the source of information becomes sparse and partially observable, more exploratory strategies such as infotaxis have been shown to be viable for finding the desired location \citep{vergassola2007infotaxis}. In the infotaxis technique, the searcher chooses an action (direction) that locally maximizes the expected rate of information acquisition, such as new sources of odor. 

In the context of visual search, the state-of-the-art general search algorithms are more oriented towards the decision-making process \citep{lanillos2013mts}. The majority of these methods are based on simplifying the sensor model as a detection/non-detection distribution \citep{Bourgault2003}. Here, the common approach is to model the sensors as a non-detection density function that depends on the state of the robot \citep{ye1999sensor,lanillos2014complex}. The drawback of such strategies is that their optimal implementation for the constrained cases\footnote{The robot has restrictions in the movement due to its kinematics \citep{Eagle1984:search_moving_target_constrained}.} are intractable \citep{Trummel1986,ye2001complexity} and are only applicable to real-time tasks in a limited domain \citep{Gan2010}.  

In the literature there are strategies that attempt to deal with the intractability of optimal search by relying on context abstraction \citep{aydemir2013active,chen2013visual}. These approaches commonly use techniques such as semantic mapping or reducing the action state space, however, they do not exploit information regarding the environment conditions that are captured by sensors. 

In visual search, sensory input is a valuable source of information and can be used (e.g. in the form of visual feature saliency) to drive the robot towards the target, in the same way as the smell of a pancake drives our non-Brownian walk on a Sunday afternoon. These salient features, which can be provided by an attention process, offer a variety of new partial information that should be incorporated into the robot decision-making algorithm.

There are a number of works that have attempted to solve the search problem by incorporating attention cues. For instance, Frintrop uses an attention framework called VOCUS \citep{frintrop2006vocus} for object detection. This computational attention system uses bottom-up saliency to generate hypotheses for possible locations of the object and then applies a classifier to the identified regions to confirm the presence of the object. Although the system shows improved detection results, it is only applied to still images and does not connect the attention mechanism to viewpoint control. In \citep{shubina2010visual}, the authors use attention in the form of viewpoint control by using a greedy algorithm to select the next angle of view and the next position to move using two different utility functions. In a more recent work \citep{Rasouli2014vsavs}, the authors take advantage of saliency cues to improve viewpoint control by directing the attention of the robot to more relevant locations. Here, the attentive capabilities, however, are only evaluated in a reactive greedy framework and the effect of using different attention schemes in the context of visual search has not been investigated. 

\subsection{Motivation}

It is anticipated that active search robots, through the explicit use of visual cues, would be able to robustly find the pursued object. However, applying such a strategy to complex vision is not well studied in both reactive and cognitive visual search approaches. For instance, semantic search accounts for contextual knowledge but does not guide the agent to the current visual stimuli. On the other hand, reactive approaches lack the capacity to use previously acquired knowledge. These shortcomings in visual search strategies point to the need for incorporating a form of attention into visual search. In this paper, we describe active visual search as a cognitive algorithm where the robot attends to relevant sensory stimuli while searching for an object and also uses past information to explore the environment. In this sense, an ideal active visual search model must be:

\begin{enumerate}[leftmargin=*]
\item \textit{Responsive}. It should be reactive to the sensory cues to ensure adaptability and generalization to any environment. 

\item \textit{Directive}. It should provide a positive directional response to stimuli that share characteristics with the sought object.

\item \textit{Spatiotemporal}. Decisions should be made by incorporating past information and spatial cues.

\item \textit{Efficient}. The time to find the object must be minimized.

\end{enumerate}

\subsection{Contribution}
We propose a new model that improves the existing works on active visual search by addressing these ideal characteristics. Our approach embeds visual attention (\textit{responsiveness} and \textit{directiveness}) in a n-step decision-making algorithm formalized as a 1st-order Markov process (\textit{spatiotemporal}). Furthermore, we show that increasing the robot's awareness of the environment  via the use of visual attention, significantly improves  performance (\textit{efficiency}). The advantage of using this approach over the existing methods is threefold: (1) all relevant visual information is directly connected to the controller, (2) action optimization is non-myopic and (3) it leverages spatial and appearance information about the object.

Figure \ref{fig:abstract} depicts the proposed solution. We encode the visual information into the belief distribution of the object location by fusing 3D information and an extension of the AIM saliency model \citep{Bruce2009saliency} for top-down color modulation by means of a backprojection procedure. The object location distribution is then generated by including the current attention stimuli and the observed measurements. The control actions are optimized to maximize the probability of detecting the target in this modified belief. 

We tested the proposed algorithm using a simulated environment because evaluating active visual search algorithms requires the ability to select and acquire images on demand for analysis. Such characteristics impose a burden for collecting and maintaining practical datasets as it requires specialized hardware, which can be very costly. As a result, to date, there is no publicly available dataset with real images that can be used for active vision applications \citep{bajcsy2016revisiting}.

Alternatively, one can use practical robotic platforms in real environments. Although deemed effective, using such practical approaches is prohibitive, both in terms of the cost and time of operation, for extensive evaluations where one may want to manipulate the environment, change the visual conditions (e.g. lighting), vary the size and structure of the environment (e.g. a regular vs disaster environment), and conduct a large number of trials.

Given such limitations, in the computer vision community, realistic simulated environments are widely used by researchers as testbeds for comparing and evaluating active vision algorithms \citep{bajcsy2016revisiting}. In our work, we also rely on a similar approach, for which we generated a large dataset of synthetic objects (e.g. household and office items) of various shapes and colors (Fig. \ref{fig:samples}) as well as an environment that resembles the interior of an office building (Fig. \ref{fig:exp1_scene}). Attempts have been made to design the simulated entities to be as realistic as possible, hence, we used professionally rendered 3D models from various online sources \footnote{The models can be found at \url{http://data.nvision2.eecs.yorku.ca/3DGEMS/}}.

The paper is structured as follows: section \ref{sec:related} describes the active visual search and its relation to passive visual search, and reviews some of the current attention models; section \ref{sec:proposed} presents the proposed attention-based active visual search; section \ref{sec:results} shows the experimental validation and results; and, finally, section \ref{sec:conclusion} summarizes findings derived from the experiments.

\section{Related work}
\label{sec:related}
\subsection{Visual search: active vs. passive}
\label{sec:search}

Visual search is defined as a perceptual task where a target (object or feature) has to be located among distractors in the environment \citep{treisman1980feature}. Strategies used for conducting visual search can either be passive or active. In the former approach processing of sensory input (images) is done via a preprogrammed set of rules and procedures. This means that the acquisition of new sensory input does not alter the way the algorithm searches for the target. An active visual search method, however, is a dynamic process in which the search strategies may change at any time depending, for instance, on new observations \citep{tsotsos1992relative}. To this end, an active approach  generalizes the search task as an optimization problem \citep{shubina2010visual} where the agent computes the best actions to find an object in the scene using the available information from the sensors.

Active optimization can take place both covertly and overtly. In a covert strategy, the algorithm decides how to analyze the 2D image of the environment based on a data or task dependent preprocessing, e.g. by using a saliency map. On the other hand, an overt approach purposefully controls the data acquisition process by either passive manipulations of camera parameters, such as focus, zoom, aperture \citep{bajcsy1988active}, or by explicit movement of the sensor in space, e.g. changing pan and tilt angles of the camera \citep{shubina2010visual}.    

In the context of visual search, one of the challenges is its intractability as the problem belongs to the NP-hard set \citep{Trummel1986,tsotsos1989complexity}. A search process can be expressed as either Minimum Time Search (MTS) and or Maximum Probability Search (MPS), is a Partially Observable Markov Decision Process (POMDP) \citep{Eagle1984:search_moving_target_constrained,lanillos2013mts}, since the agent observes only a portion of the environment reached by the sensors. Denoting the time to detect the object as a random variable $T$, the problem is defined as finding the actions that minimize 
\begin{align}
 \min E\{T\} = \min \int_{k=1}^\infty (1- P(T\leq k)).
 \end{align}
 
 
Currently, the best approaches in the literature for real-time search applications are open loop n-step controllers \citep{lanillos2013mts}. Here, the difficulty lies in obtaining a good estimator that computes the expected time to detect the object after performing $n$ actions. The easiest solution is to compute 1-step greedy approximations \citep{Bourgault2003}. Other authors semantically abstract the environment \citep{aydemir2013active} and plan on a higher level. A different approach, which only works in the MPS, is to exploit the kernel properties of the cumulative probability function \citep{tseng2015near} assuming smooth and continuous distributions.  Another interesting approach has been shown in touch active learning \citep{kaboli2017activetouch}, where proximity sensors provide the relevant cues for searching the object. Nonetheless, these methods do not take into account all the useful information provided by the sensors except the binary detection/non-detection response.
\vspace*{-\baselineskip} 
\subsection{Saliency as a form of visual attention}
\label{sec:attention}
Visual saliency, as one of the underlying representations that supports attentive processes, has been extensively studied. Two classes of algorithms are used to construct visual saliency maps \citep{bylinskii2015towards}:

1- \textit{Bottom-up} approaches which are data driven and measure saliency based on detecting the regions of the image that stand out in comparison to the rest of the scene. Bottom-up algorithms are categorized into two groups of \textit{object-based} or \textit{fixation-based} (or \textit{space-based}) saliency.

As the name implies, object-based saliency models are concerned with finding the extent of salient objects in the image. These algorithms, in essence, are similar to segmentation methods. However, instead of partitioning the image into regions of coherent properties, object-based models highlight objects that stand out the most in the image. As a preprocessing stage, these algorithms often rely on over-segmentation methods to divide the image into smaller regions in the form of equal patch sizes \citep{Goferman2012} or superpixels \citep{Chang2011fusing,Jiang2011salobj}. The object-based saliency methods find closed contour regions that resemble an object. This means they are effective when there is a clear boundary between the objects and the background. 

The fixation-based models predict human eye fixations typically measured by subjective rankings of interesting and salient locations or eye movement \citep{Borji2013quantanalysis}. In contrast to the object-based methods, these models identify saliency at the pixel-level by exploiting different types of features ranging from simple low-level ones such as color or intensity \citep{Itti1998rapidscene} to higher level learned features generated by methods such as sparse coding \citep{Hou2008dva} or Independent Component Analysis (ICA) \citep{Bruce2007aim}. The distribution of these features is measured either locally or globally to identify the uniqueness of a given region in the image. In more recent works, machine learning techniques, such as deep learning and neural nets, are also used to predict bottom-up saliency \citep{Li2015cnnsal,Zhao2015deepsal}.

2- \textit{Top-down} saliency models, as opposed to the bottom-up approaches, identify saliency as the regions with similar properties to a specific object or task \citep{Cave1999top}. Many of these models treat saliency as a classification problem using high-level features such as SIFT in conjunction with learning techniques such as SVM \citep{Zhu2014context}, conditional random fields (CRF) \citep{JYang2012tpdown} or neural networks \citep{He2016deeptop} to determine the presence of the object of interest based on a combination of pre-learned features.
\vspace*{-0.2cm} 
\subsection{Visual attention in robotics}
\label{sec:rob_attention}
Visual saliency models have been increasingly used in robotics for applications such as obtaining robust and salient features for 3D mapping and localization \citep{Kim2013slamsal}, object recognition \citep{Orabona2005,frintrop2006vocus}, and high-speed navigation in cluttered environments \citep{Roberts2012}. Social robotics is also benefiting from the use of visual saliency. For instance, in \citep{N.J.Butko2008} and \citep{Moren2008bioatt} top-down saliency is employed to identify humans based on detected motion patterns. Visual saliency is also used to estimate human gaze in HRI applications \citep{A.P.Shon2005gaze}, in task understanding for environment manipulation \citep{Ude2005disatt} and for grasping objects \citep{Kragic2005grasp}. In addition, more general attention systems are investigated in robotics \citep{ferreira2014attentional}. For instance, in \citep{lanillos2015designing} an attention mechanism has been used as a core middleware for achieving correct social behavioral responses.

In the context of visual object search, the use of visual saliency improves the performance by prioritizing the search regions to maximize the chance of finding the target \citep{Rasouli2014vs,Rasouli2014vsavs}. In these works, however, the benefits of different saliency approaches are not examined in isolation, the object chosen for the experiments is rather simple and not representative of common everyday objects. Furthermore, the search algorithm is evaluated only in structured environments, thus failing to demonstrate how the increased clutter in highly unstructured environments can impact the efficiency of search. The optimization in these works relies only on a one-step look-ahead greedy approach leaving the question of how such models can be incorporated into non-myopic algorithms open-ended.

\section{Attention-based active search}
\label{sec:proposed}

This work addresses an extreme unconstrained version of overt visual attention since the robotic platform can move freely in the scene. The challenge is not only to determine the visually salient stimuli, but also to generate a set of control actions that will lead the robot to find an object in a completely new environment.

\subsection{A model for active visual search}

We define the visual search strategy as the fusion of information gained via new visual stimuli combined with the knowledge accumulated from past observations. In this approach, the search starts by capturing an image of the environment and processing it by the recognition algorithm. If the target is detected, the search process terminates, otherwise, using an attention model, we identify potential locations where the object of interest might be. We combine this information with the past knowledge exploitation term, based on the theory of optimal search, to achieve exploratory behavior. Furthermore, as the informative cues could come from different sources that may not relate to the target, we introduce an inhibitory term that reduces the relevance of a specific stimulus-location if the object is not found. 

Throughout this paper, we will use the following notation for our problem statement and formulation:\\

\noindent $b^k$\quad -\quad Object location belief/distribution

\noindent $s^k$\quad -\quad Mobile robot state (pose) at instant $k$

\noindent $u^k$\quad -\quad Robot action at instant $k$

\noindent $\tau^k$\quad -\quad Object state 

\noindent $z^k$\quad -\quad Observation measurement

\noindent $f(k)$\quad -\quad  Saliency cue

\noindent $P(.)$\quad -\quad  Probability distribution

\noindent $D / \overline{D}$\quad-\quad{}Detection / non-detection event\\

Conceptually, the proposed attention-based active visual search strategy is a trade-off between the probability of detecting the object given the relevant visual cues modulated by the inhibition, and the probability of detecting the object given the past observations. Using this formulation, the next location to search can be obtained by:

\begin{align}
\max P(z^{k+1} = D, \tau| z^{0:k}, s^{0:k})
\end{align}

\noindent where the probability of the target location $P(\tau)$ at instant $k$ is defined by the fused information:

\small
\begin{align}
\underbrace{\epsilon P(\overbrace{\tau}^\text{object}| \overbrace{f(k)}^\text{Stimuli}, s^{k})}_\text{New stimuli exploitation} & \!\overbrace{\mathcal{I}}^\text{Inhibition} + \nonumber\\  + &\underbrace{(1-\epsilon)P(\tau| \overbrace{z^{0:k-1}\!=\!\overline{D}}^\text{Observations}, s^{0:k-1})}_\text{Past stimuli exploitation}
\end{align}
\label{eq:search}
\normalsize
where $\epsilon$ is a parameter that determines the influence of the new stimuli. We set the value of $\epsilon$ empirically. In Eq. \ref{eq:search}, the first term introduces exploitation behavior of new attention stimuli and the second term produces exploratory or exploitation actions while taking into account the past knowledge. See Section \ref{sec:proposed:belief} and \ref{sec:proposed:decision} for further details.

\subsection{System design}
\label{sec:proposed:design}
\begin{figure}[!t]
\centering
\includegraphics[width=\columnwidth]{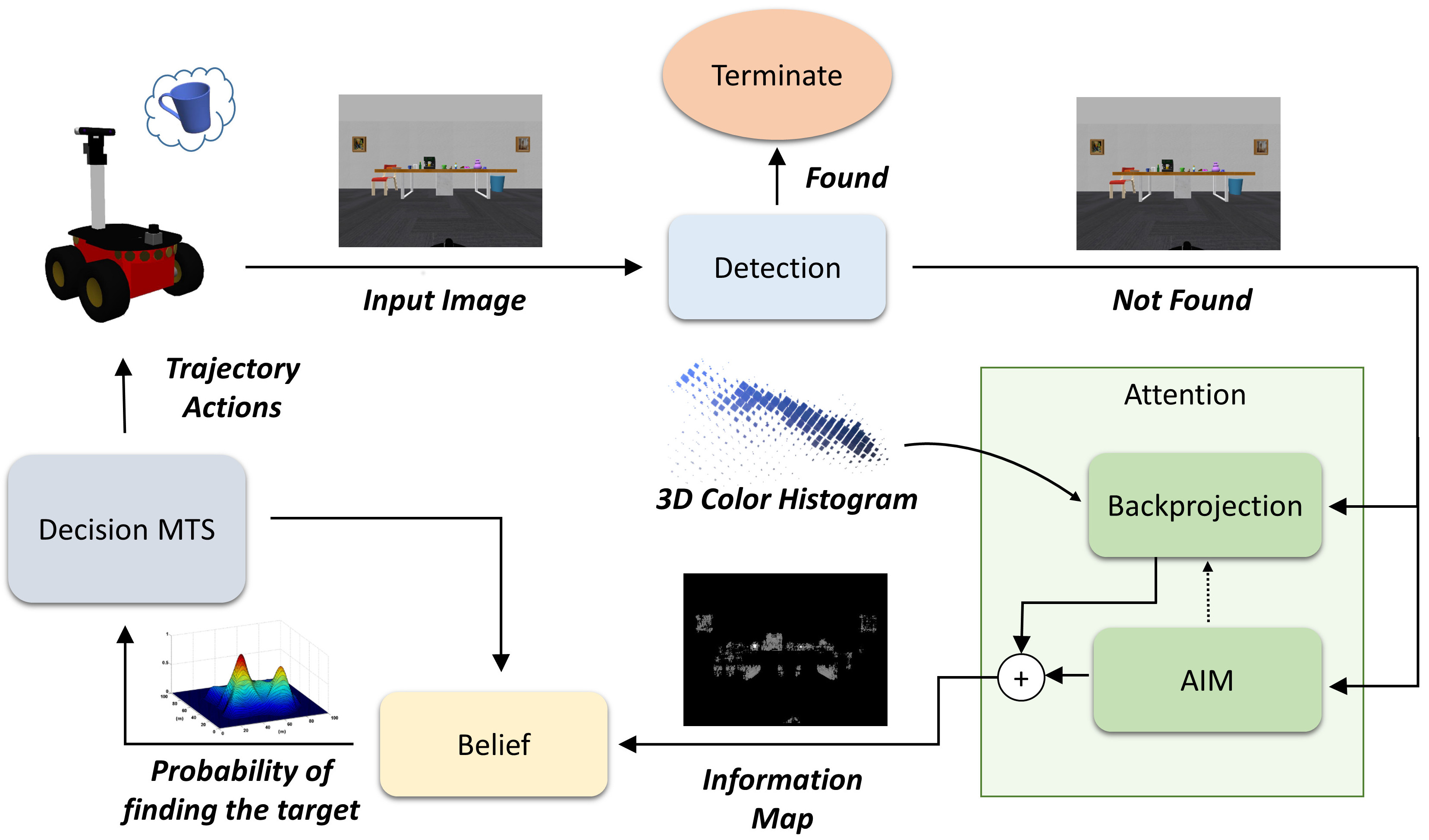} 
\caption{The overall structure of the proposed search model. The robot captures an image of the environment and then applies the recognition algorithm. If the object is found the search is terminated, and if not, the image is passed to the attention module. The attention module produces an information map which is used to update the beliefs of the robot about the potential locations for the sought object.} 
\label{fig:overall}
\end{figure}

The proposed system has four subcomponents: \textit{detection}, \textit{attention}, \textit{belief} and \textit{decision}. Fig. \ref{fig:overall} depicts the overall design of the system and the flow of control from capturing an image to making the next decision.
\newline

\vspace*{-\baselineskip} 
\subsection{Attention}
\label{sec:proposed:attention}

\begin{figure}[!hbtp]
\centering
\includegraphics[width=0.45\textwidth]{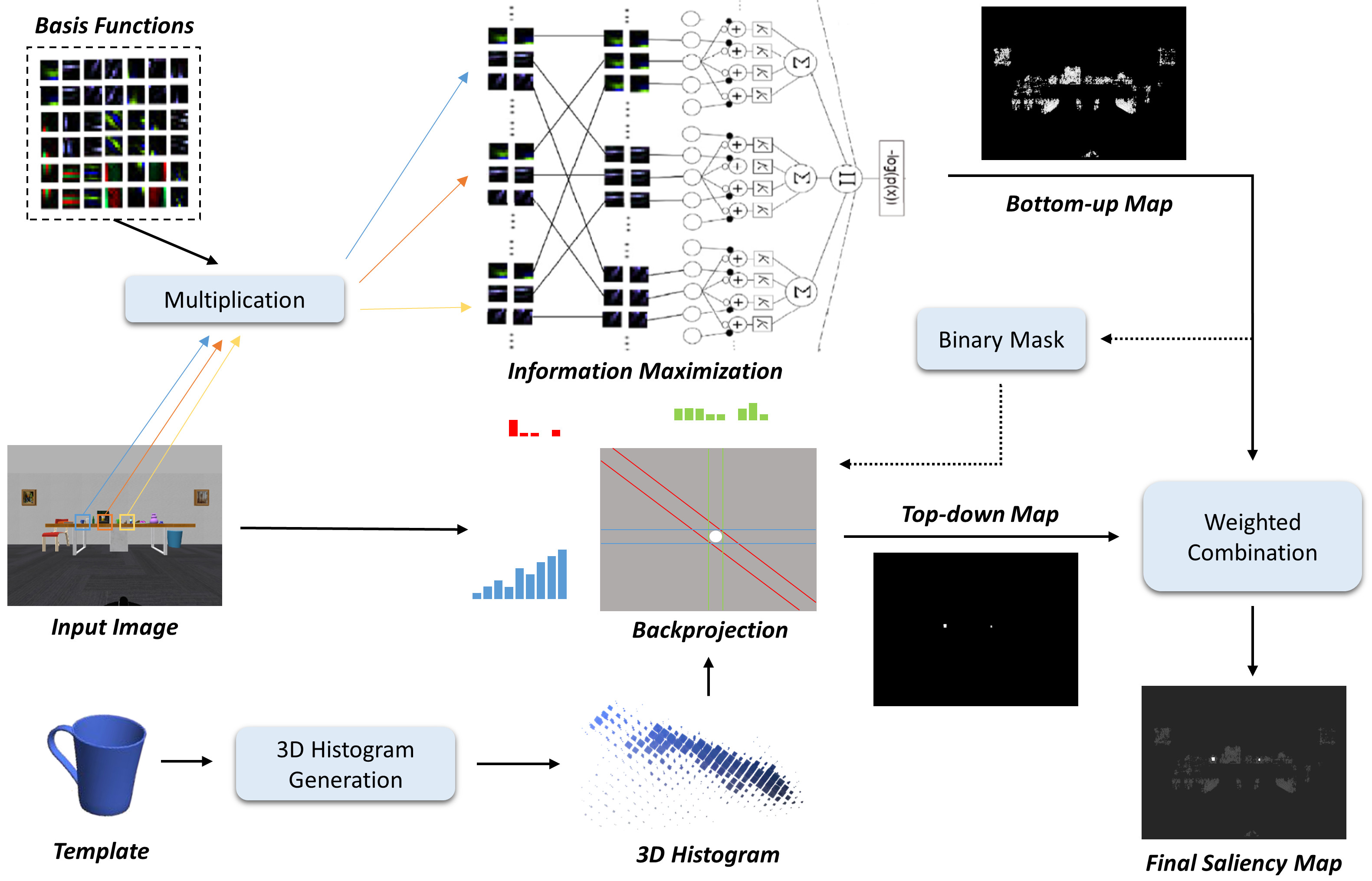} 
\caption{The overall process of generating saliency map in search. The attention module receives the input image, generates a top-down and a bottom-up saliency map and then combines them to create an information map.} 
\label{fig:attention}
\end{figure}

Fig. \ref{fig:attention} illustrates the overall structure of our attention model which is inspired by \citep{Rasouli2014vsavs}. The attention module comprises two subcomponents: a \textit{bottom-up} and a \textit{top-down} module. Depending on the configuration of the search a combination of these two modules are used.  

\subsubsection{Bottom-up model}
\label{sec:proposed:bottomup}
The bottom-up algorithm is used to identify interest regions that might have some form of spatial relation to the object of interest. These regions can be surfaces such as tabletops, shelves, chairs, or, in the context of search and rescue scenarios, areas with debris, broken structures or furniture. 

The search environment is often very cluttered, hence, the bottom-up algorithm should be able to perform some form of ranking to distinguish between a common redundant structure and a rare one. For this purpose, we use the Attention based on Information Maximization (AIM) algorithm \citep{Bruce2007aim}. 

The AIM algorithm is based on the information maximization technique allowing this model to identify regions that yield the most information, i.e. the areas that are more unique. In the context of search and rescue, this is particularly useful because this measure of uniqueness not only corresponds to the presence of a rare feature but also to the absence of features in the scene, e.g. a hole in a wall.

The AIM algorithm uses Independent Component Analysis (ICA) \citep{D.Langli2010} generated features. To obtain these features, we first train our ICA model over a large sample of images collected from areas similar to our search experiments, e.g. office environments, furniture, etc. The result of this training is a set of basis functions. 

The AIM algorithm starts by convolving the ICA generated basis functions with the input image. Then it computes the joint likelihood of the responses using a Gaussian window. The overall probability density function of features is computed as the product of each individual probability assuming the independence of the ICA features. At the end, the self-information measure of each distribution is calculated as a sum of negative log-probabilities of each filter response. The higher the value of self-information measure at each point, the rarer it is within the image and therefore is recognized as salient.

The resulting information map from AIM is thresholded by some percentile value $th_{AIM}$ (which is set empirically) to find the maximum responses.

\subsubsection{Top-down model}
\label{sec:proposed:topdown}

As pointed out earlier in \ref{sec:attention}, top-down attention models often treat saliency as a classification problem similar to those used in recognition applications. Such methods rely on high-level features which require a certain amount of visibility to be detectable.

Nevertheless, in the context of visual search, we use the top-down model to obtain clues regarding the target's presence in regions that are far away from the camera and are undetectable by the recognition algorithm. Therefore, we use color as the feature of choice because it is view-invariant and can be easily detected from the distance. 

To detect color similarities in the scene, we use the histogram backprojection (BP) technique \citep{Swain1991}. A BP map is generated as follows: let $h(C)$ be the histogram function which maps colorspace $C = (a_1,a_2,...,a_i)$, where $a_i$ is the $i^{th}$ channel of $C$, to a bin of histogram $H(C)$ computed from the object\textquotesingle s template, $T_\Theta$. The backprojection of the object's color over an image $I$ is given by,  

\begin{align}
\forall x,y : b_{x,y} := h(I_{x,y,c})
\end{align}

\noindent where $b$ is the grayscale backprojection image.

\subsubsection*{Visual saliency map}
In this work, we incorporate the visual saliency results in four different ways. The first two models are using each method of saliency in isolation. Throughout this paper we will refer to bottom-up and top-down saliencies as  \textit{BU} and \textit{TD} respectively.

The third model is a weighted combination of both methods given by, 
\begin{equation}
\begin{split}
f(k) & = \eta \omega _{a}info^\prime (k) + \omega_{b} bp(k)\\
& \omega_a + \omega_b = 1
\end{split}
\end{equation}
where $\eta$ is the normalization parameter, $info^\prime(k)$ is thresholded information map, $bp(k)$ is the backprojection map, and  $\omega_a$ and $\omega_b$ are the weights of AIM and BP respectively. We call this method \textit{BU+TD}.

The fourth attention model is similar to \textit{BU+TD} but instead of directly applying BP to the color image, we preprocess the image as follows: we first generate a binary map from  $info^\prime (k)$ and then pixel-wise multiply it with the color image,

\begin{equation}
\label{eq:salmap}
\begin{split}
& Im^\prime (k) = Im(k)\odot M(k)\\
& \Bigg\{
\begin{tabular}{c c}
$M(k) = 1$ & $info^\prime (k) > 0$\\
$M(k) = 0$ & otherwise
\end{tabular}
\end{split}
\end{equation}

\noindent where $M(k)$ is the binary mask, $Im^\prime(k)$ and $Im(k)$ are the filtered color image and the image respectively and $\odot$ is pixel-wise multiplication operator. Then the BP algorithm is applied to the  filtered color image to generate top-down saliency which is then linearly combined with $info^\prime (k)$ as in Eq.\ref{eq:salmap}. Therefore, using this method we only generate top-down saliency from the regions identified as important by the bottom-up method. We refer to this model as \textit{BU+BU$\odot$TD}.

\subsection{Camera sensor model}
\label{sec:proposed:camera}
The decision-making algorithm uses a camera model designed as a non-detection density function that depends on the robot/camera and object pose: $P(z^k=\overline{D}|s^k,\tau^k)$. For this purpose, the state of the camera is defined as the 2D location and the angle in the horizontal axis ($s^k = (x,y,\phi)$). Thus, we use a distance model $\mathcal{D}$ that decreases exponentially as the target goes further away \citep{Gan2010}. In addition, an angle model $\mathcal{A}$ is used to define the field of view of the camera. Here, the probability of detecting the object decreases as the view angle deviates from the axis defined by the principal point and the vehicle yaw $\phi$.

\noindent The distance non-detection density function is:
\begin{align}
\mathcal{D} = \exp\left(\frac{-\sigma}{d_{max}^2} ||s^k-\tau^k||^2\right)
\end{align}
\noindent The angle non-detection density function is formulated as:
\begin{align}
\mathcal{A} = \frac{\beta}{ 2\alpha\Gamma(\frac{1}{\beta}) } \exp\left(-\frac{|\phi|^\beta}{\alpha^\beta}\right)
\end{align}
Finally, we define the camera model by combining the angle and the distance functions, 

\begin{align}
\label{eq:sensor}
P(z^k=\overline{D}|s^k,\tau^k) = 1 - \left[P_{d_{max}}\cdot\mathcal{D}\cdot\mathcal{A}\right]
\end{align}
where $\alpha, \beta, d_{max}$ and $P_{d_{max}} \in [0,1])$ are parameters that should be tuned depending on the range and detector model used. For instance, $P_{d_{max}} = 0.8$ means that the detector has a confidence of $0.8$ for a positive detection.

\subsection{Belief construction}
\label{sec:proposed:belief}
We first construct the 2D projection of the visual saliency model by summing up the columns of the 3D occupancy data. This projection becomes the belief map given the visual cues $P(\tau| f(k))$. 

Then, the object location belief $b^k$ is obtained by fusing projected visual saliency and an inhibition distribution generated by the past observations $\mathcal{I}^k = P(\tau, z^{0:k} = \overline{D} | s^k)$.
\begin{align}
 b^k &= \eta P(\tau| s^k, z^k)  \propto P(\tau| f(k)) \mathcal{I}^k
\end{align}
where $\eta$ is a normalization parameter that makes $\sum b^k = 1$ and $f(k)$ is the saliency signal. The last term accounts for a spatial memory that acts as an inhibitory signal ($\mathcal{I}^k$) and ensures that a place that has been well observed is not investigated any further. The inhibition map is modeled as a geometric monotone decreasing function. Its value decreases as the distance to the stimuli gets smaller. The function in recursive form is:
\small
\begin{equation}
\mathcal{I}^k =  \frac{1}{2}\frac{\lVert s^k - f(k)\rVert}{(d_{max}-d_{min})} \mathcal{I}^{k-1} \quad \text{if } f(k) > 0
\end{equation}
\normalsize
where $d_{max}$ and $d_{min}$ are maximum and minimum observed distances.

In order to get the final object location distribution, we fuse the current generated belief with the past knowledge by performing a smoothing operation controlled by the parameter $\epsilon$, which adjusts the level of influence of the newly observed cues (i.e. how reactive is the robot to new relevant stimuli).
\begin{align}
 b^k \propto \epsilon b^{k-1} + (1-\epsilon)b^k
\end{align}

Regardless of the presence of relevant stimuli, the belief of the target location is updated recursively for static objects using the sensor model equation (Eq.\ref{eq:sensor}) to include the negative observations into the belief of the robot:
\begin{align}
b^{k} = \eta P(z^{k} = \overline{D} | s^{k}, \tau^k) b^{k-1}
\end{align}

\subsection{Decision-making}
\label{sec:proposed:decision}

The actions are optimized by maximizing the probability of detecting the target given the attention cues encoded in the current belief of the system. The robot will select the actions that maximize the information obtained by the sensors according to the belief about the target location. The general optimization function is given by:

\begin{align}
u^* = \arg\max_{u} \mathop{{}\mathbb{E}}\left[ P(z^{k:\infty} = D, \tau | b^{k:\infty}, s^{k:\infty}) \right]
\end{align}
where the actions are computed to maximize the probability of detecting the object in the environment. Note that we are assuming that there is a function that transforms the action $u$ into the state of the robot $s$.

To make the optimization tractable, we approximate the expectation by introducing a heuristic as proposed in \citep{lanillos2014:MTScontinous}. Afterwards, we integrate over the target location variable $\tau$, since it is an unknown variable. Under the 1st-order Markovian assumption and assuming that the observations events are independent in different instants, the probability of detecting the target can be simplified to:

\small
\begin{align}
&P(z^{k:\infty}  = D, \tau | b^{k:\infty}, s^{k:\infty}) \approx \nonumber\\
 &\approx 1 - \int_{\tau} P(z^{k:k+n} = \overline{D} | s^{k:k+N}, \tau) H(b^{k+n+1}, s^{k+n+1}) b^k d\tau\\
 &= 1 - \int_{\tau} \prod_k^n P(z^{k+i} = \overline{D} | s^{k+i}, \tau) H(b^{k+n+1}, s^{k+n+1}) b^k d\tau
 \label{eq:utility}
\end{align}
\normalsize

\noindent where the $H(.)$ term is the heuristic function that approximates the future observations and depends on the last state and belief of the optimized trajectory ($k+N+1$). The heuristic, which is modeled as a distribution, is as follows,
\begin{align}
H(b^{k+n+1}, s^{k+n+1}) = 1- \lambda^{\frac{\parallel s^k-\tau \parallel}{V}}.
\end{align}
Here, $\lambda$ is a discounted time factor ($0 \leq \lambda \leq 1$) and $V$ is the velocity of the robot.
Note that, internally, the decision-making algorithm computes the predicted observations ($k:k+n$) using the sensor model in Eq. \ref{eq:sensor} and the dynamic model of the robot. That is, we can evaluate in a forward manner any trajectory of the robot.

Once we have the utility function (Eq. \ref{eq:utility}), the optimal piecewise actions $u^*$ for the horizon $k$ to $k+N$ can be computed using the interior-point algorithm\footnote{Although the optimization can be computed using a gradient-based approach due to the properties of the belief, with this algorithm we can also tackle some degenerate cases where non-linearities appear.} with finite differences to approximate the gradients. For computing the explicit gradient please refer to \citep{Gan2010}.

\section{Results}
\label{sec:results}

\subsection{Experimental design}
\label{sec:experimental}
\begin{figure}[!hbtp]
\centering
\includegraphics[width=0.8\columnwidth]{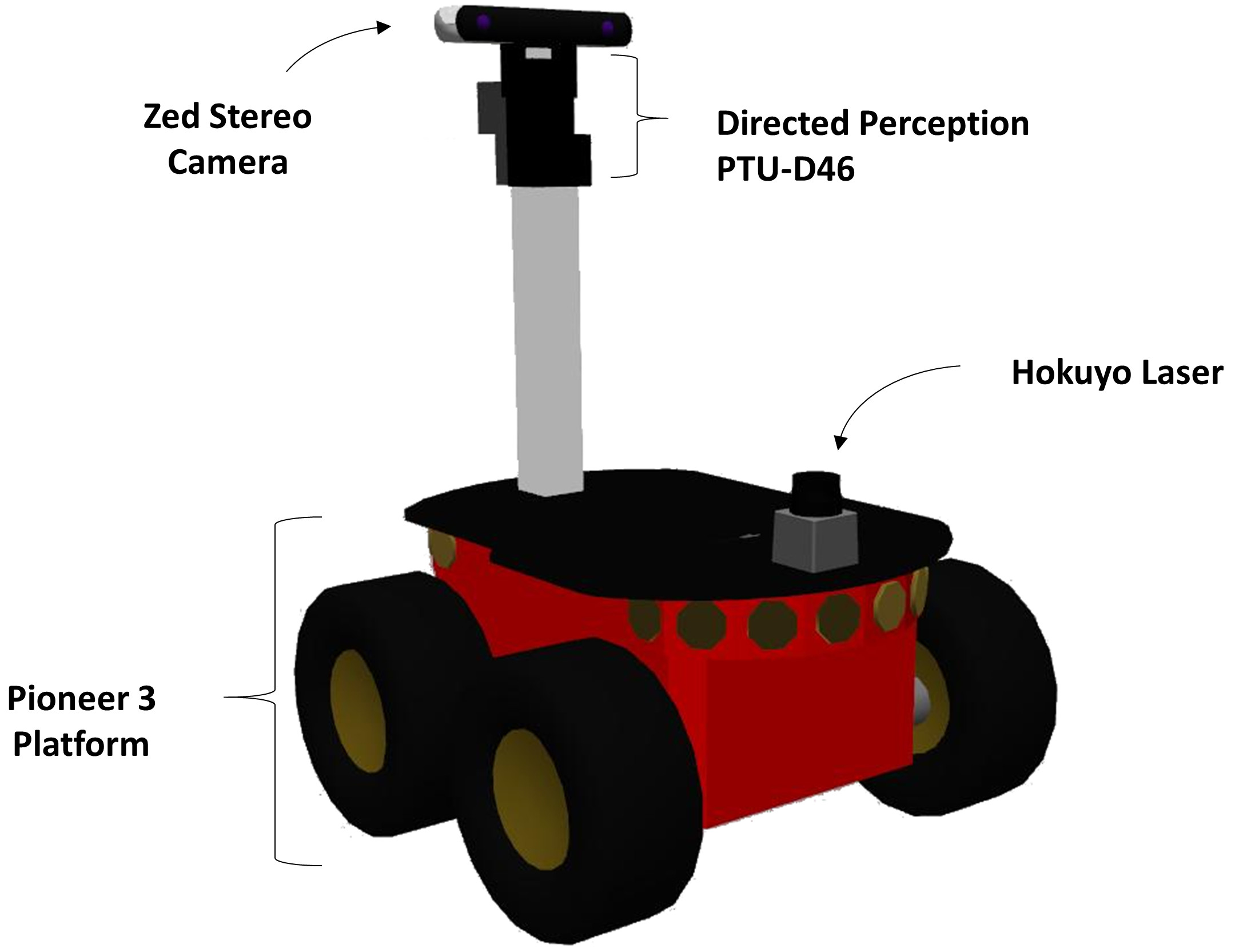} 
\caption{A view of the pioneer robot which was simulated in Gazebo with its integrated sensors.} 
\label{fig:searchbot}
\end{figure} 

The robot 3D model is based on Pioneer 3, a skid-steer 4-wheel mobile platform. It is equipped with a Hokuyo Lidar scanner for navigation, and a Zed stereo camera mounted on a pan-tilt unit (which is fixed in our experiments). The experiments are conducted in a simulated environment using the Gazebo simulation software \citep{Koenig2004}. We designed a large number of Gazebo 3D models of household and office items for the experiments. Fig. \ref{fig:samples} shows some of the objects used in the experiments. The dimensions of the search environment in all experiments is $20\times 20 m^2$.

All the communications with the platform were done through ROS. For navigation, we used \textit{gmapping} and \textit{move base} packages implemented in ROS. The maps were created using the simulated Lidar sensor and were generated on the go in each experiment, i.e. the environment configuration was unknown to the system.

As for the robot dynamics, the maximum and minimum velocities were set to $0.7$ and $0.1 m/s$ respectively. The maximum rotational velocity was set to $45^o/s$ and the acceleration limit to $2.5 m/s^2$ in $x$ and $y$ directions.
 
For the top-down saliency model, we chose the C1C2C3 colorspace which provides a color representation robust to illumination changes both in terms of detectability and discriminability of colored objects in cluttered environments \citep{rasouli2017effect}.

In the following subsections we use the search algorithm without the attention model as a baseline and refer to it as (\textit{NOSAL}). Then, we report on the results in terms of the percentage of change in the performance comparing to the baseline $\left(\frac{\text{proposed method measure}}{\text{baseline measure}}\right)$.

\begin{table}[!hbtp]
\caption{Defined parameter values for the experiments}
\centering
\resizebox{\columnwidth}{!}{
\small
\begin{tabular}{cccc}
\toprule
\multicolumn{1}{c}{\textbf{Parameter}} & \textbf{Notation} & \textbf{Value}  \\ 
\midrule				
Map size & {-}	& {$20\times 20$, $1m/node$}	\\ 
\midrule				
Distance Model & {$\sigma$,$d_{max}$ }	& {($0.4$, $3$)}	\\ 

\midrule
Angle Model	& {$\beta$, $\alpha$, $FOV$}& {($100$, $1$, $110^o$)}	\\
\midrule
Detection & {$P_{dmax}$, $d_{min}$}	& {($0.9$, $0.4$)}	\\

\midrule
AIM  & {$th_{AIM}$, $\omega_a$}			& {($0.95$, $0.2$)}	\\
\midrule
BP & {$colorspace$, $bins$, $\omega_b$}	& {($C1C2C3$, $64$, $0.8$)}	\\
\bottomrule
\end{tabular}
}
\label{table:parameters}
\end{table}
\normalsize

\subsection{Selecting optimization parameters}

In the following experiments, we set the optimization parameters, i.e. the number of steps (future actions) that the algorithm plans, and the number of actions that are actually executed. Since the search environment is unknown at the beginning, all actions planned by the algorithm may not be executable, e.g. the algorithm might select a location for the robot that is occupied by an obstacle. On the other hand, planning several actions ahead can produce a more globally optimal solution. As a result, here we want to explore the trade-off between the planning and execution.

We ran approximately 300 trials in which we put 8 targets in 3 different arrangements and in each case placed the robot in a fixed location with respect to the objects. We altered the level of planning from 2 to 4 steps ahead and in each case executed from only 1 action to the maximum number of planned ones. For instance, if 3 steps ahead were planned by the algorithm, in each round we executed 1, 2 and 3 actions.

The result of these tests was inconclusive in a sense that there is not a single configuration of the global optimization that results in the best performance in all scenarios. In fact, the performance varied depending on the configuration of the environment and the type of the search method used. Hence, we averaged the results and used the optimization parameters, 3 steps planning and 2 executions,  that resulted in a fairly good performance for all search models.
 
\subsection{Experiment 1: 3D object search in a complex environment}
\label{sec:results:1}

\begin{figure}[!t]
\centering
\subfigure[]{\includegraphics[width=0.35\textwidth]{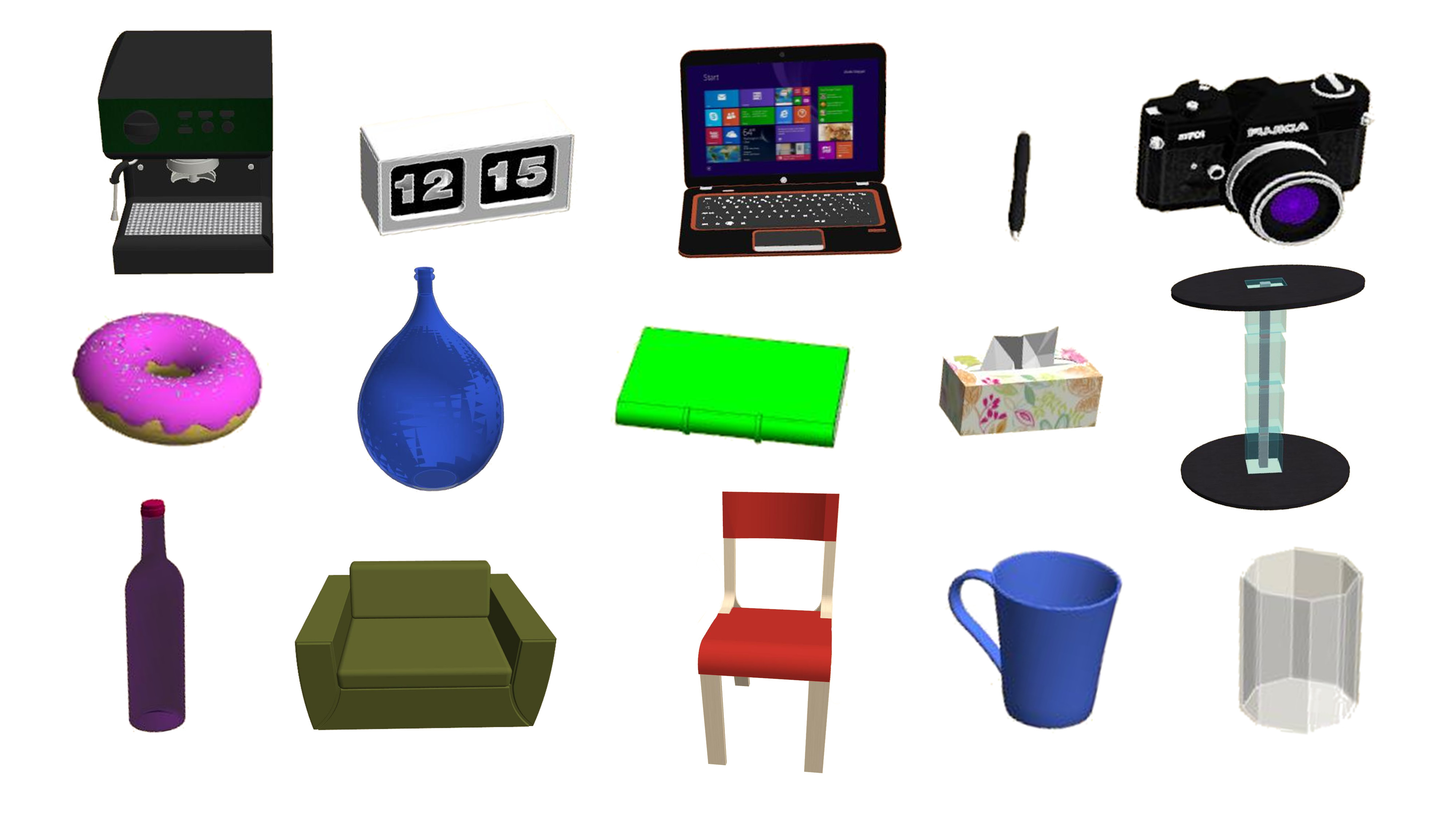}
\label{fig:gazebo_samples}}
\hfill
\subfigure[]{\includegraphics[width=0.08\textwidth]{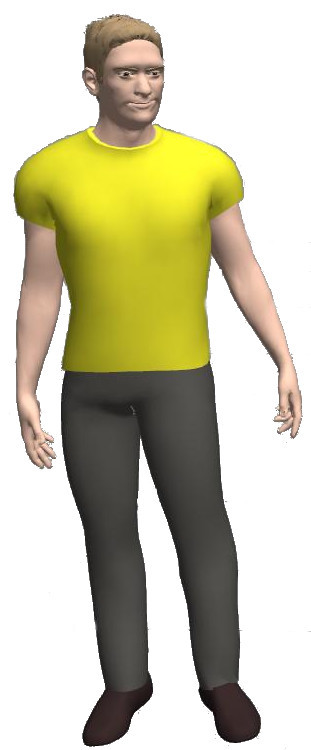}
\label{fig:human_sample}}
\caption{A sample of a) objects and b) human models used in Gazebo simulation to model a realistic environment.} 
\label{fig:samples}
\end{figure} 

In the first experiment, we examined the search algorithms in a typical environment with a complex structure. We designed a location resembling an office environment with furniture, electronics, decorations, etc. We populated the environment with a variety of objects (see Fig. \ref{fig:gazebo_samples}) to maximize background clutter in the scenes.

\begin{figure*}[!thbp]
\centering
\subfigure[A snapshot of the environment]{\includegraphics[width=0.40\textwidth, height=160px]{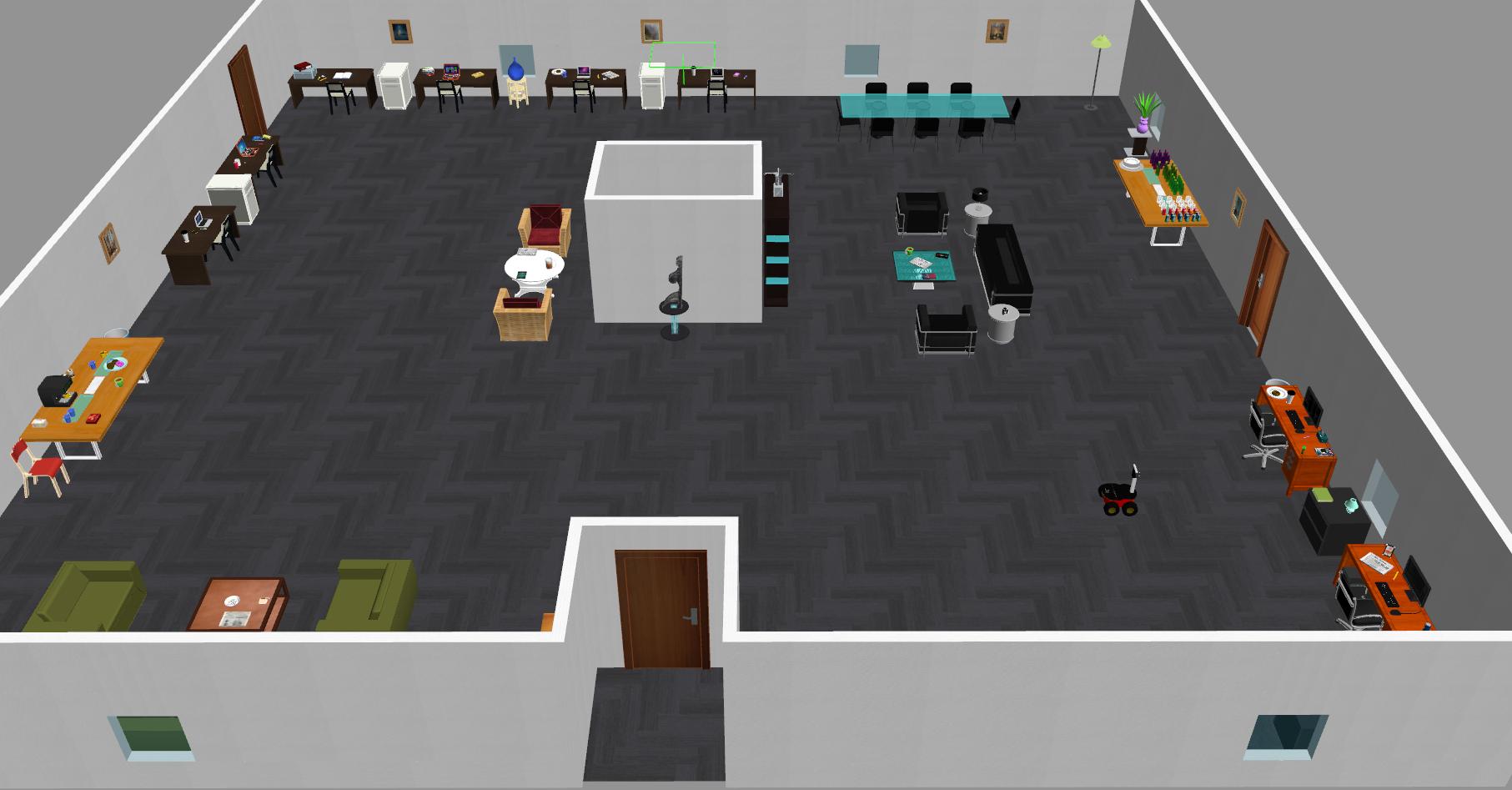} 
\label{fig:exp1_scene}}
\hspace{0.2cm}
\begin{minipage}[b]{.50\textwidth}
\centering
\subfigure[Sought objects: box, glue and Homer figurine]{\includegraphics[width=0.6\textwidth]{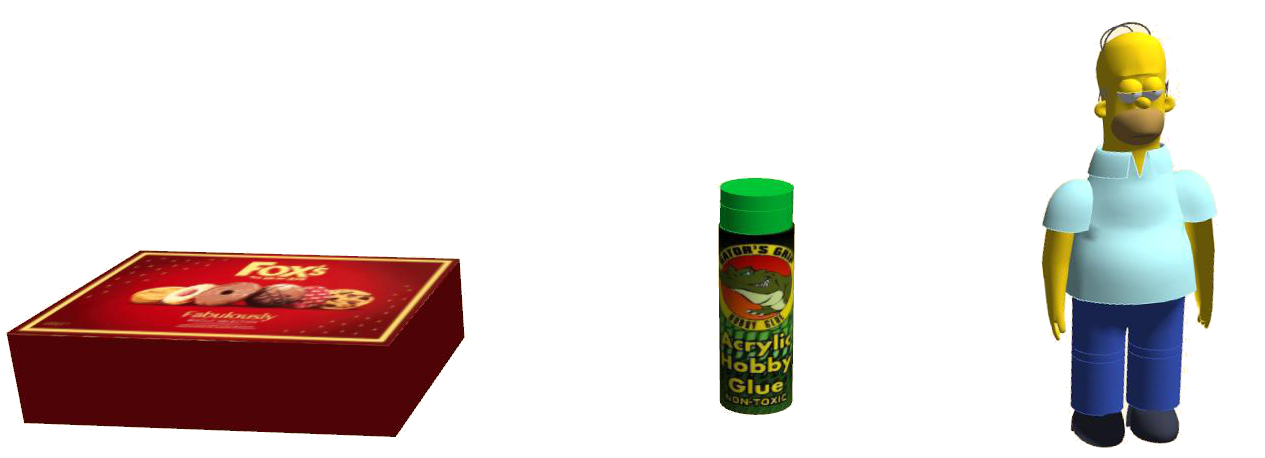}
\label{fig:exp1_objs}}
\subfigure[Mean number of actions to find all objects]{\includegraphics[width=0.45\textwidth]{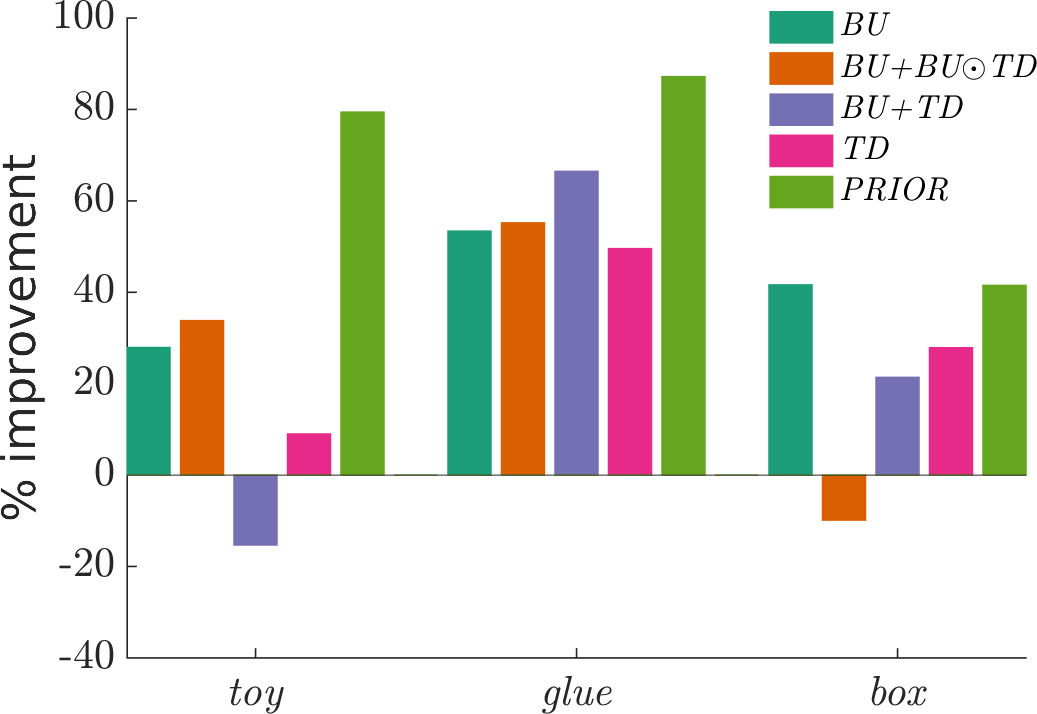}
\label{fig:exp1_acts}}
\hspace{0.2cm}
\subfigure[Mean time of the search to find all objects]{\includegraphics[width=0.45\textwidth]{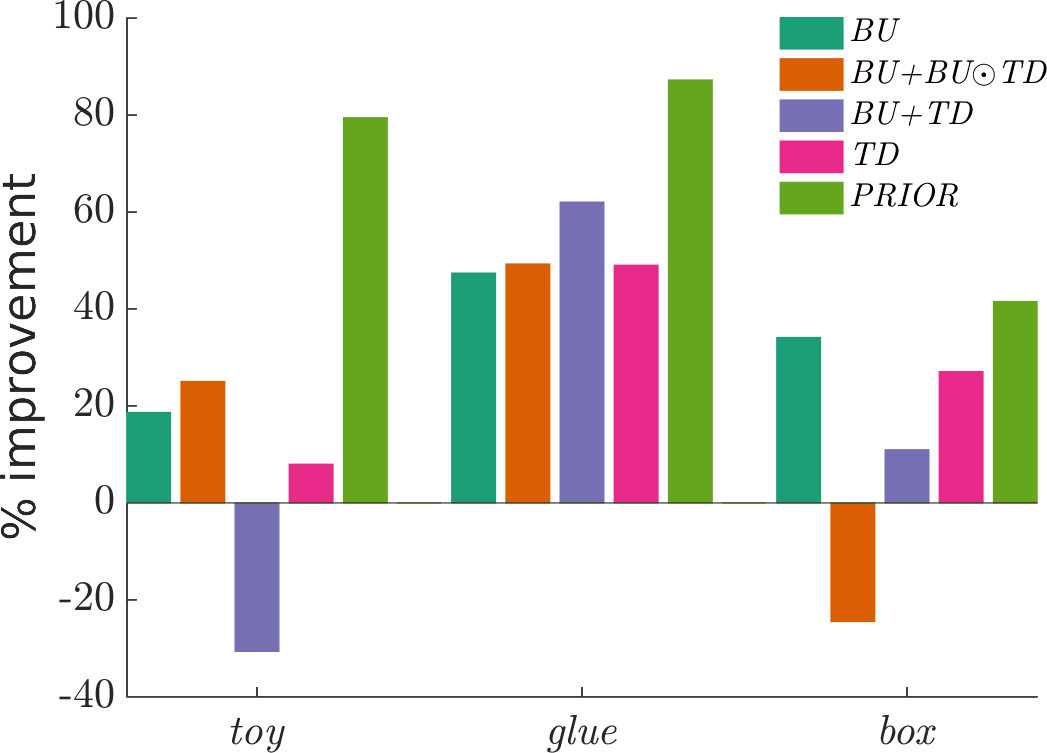}
\label{fig:exp1_time}}
\end{minipage}
\caption{Experiment 1, a cluttered office environment. One of the three objects (b) have to be found in each trial. The performance is measured in terms of (c) the number of actions needed by the agent to find the object and (d) the time of the search.}
\end{figure*}

Three objects (Fig. \ref{fig:exp1_objs}) were chosen as search targets each with different level of color similarity to the environment background. Each target has a different dominant color, red (biscuit box), green (glue bottle) and blue (the homer toy). The objects and the robot locations were randomized three times for a total of 27 search configurations. 

In this experiment, in addition to the proposed methods,  we used a method similar to \textit{NOSAL} with prior knowledge of the object's location. We modeled the prior as a normal distribution $\aleph (\mu , \sigma ^2)$ where $\sigma = 3.5$ and $\mu$ is set at the center of the object. The reason for this choice is to use the search with prior knowledge as a baseline to determine how much improvement is achievable using visual saliency.

\begin{figure*}[!h]
\centering
\includegraphics[width=1\textwidth, height=225px]{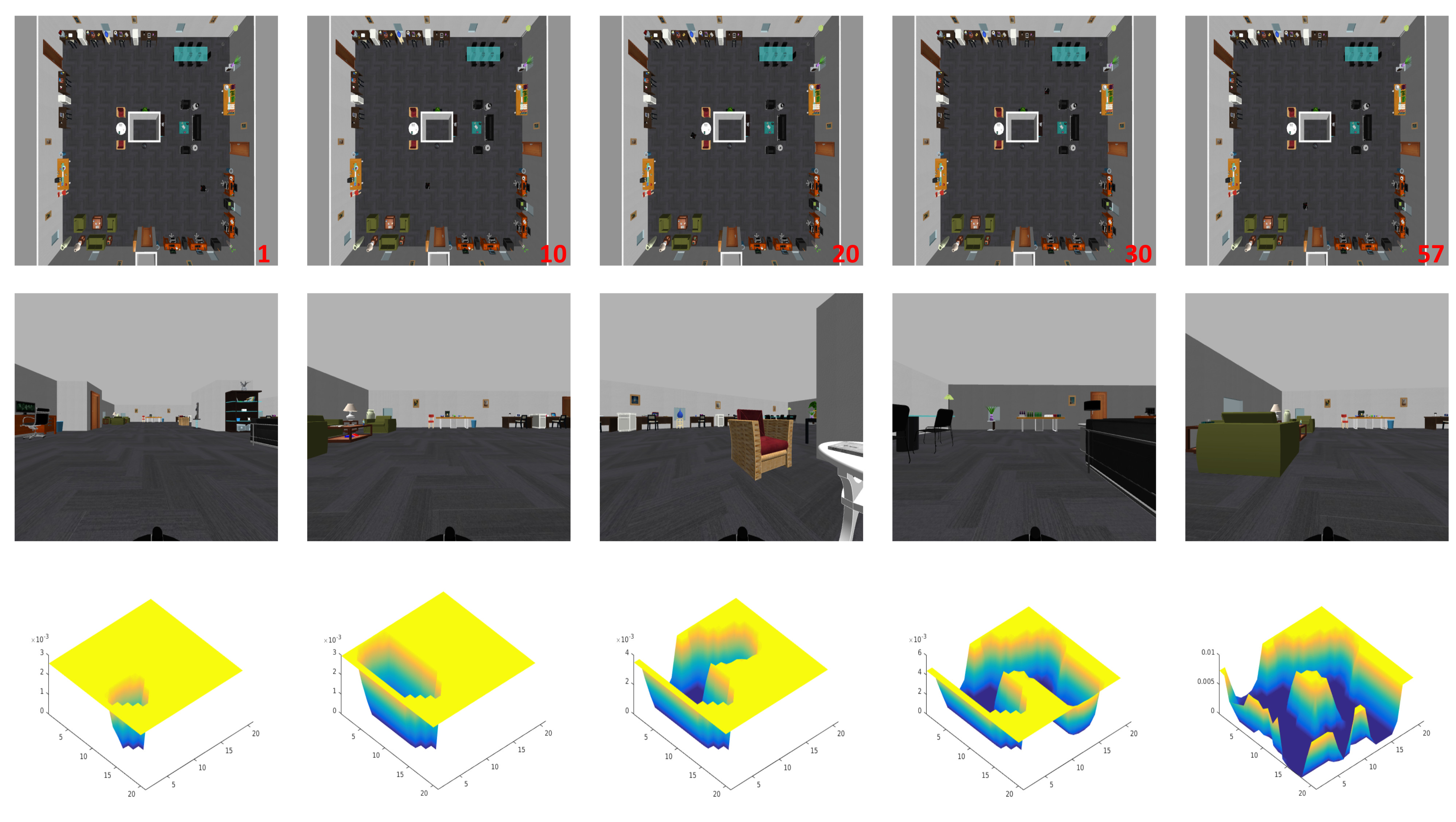} 
\caption{Experiment 1. A sample search run using the \textit{NOSAL} method. The object of interest is the Homer toy, which is placed on the coffee table between the green sofa set at the bottom-left corner in the top-down view. From the top: top-down view, robot's view and the probability of finding the target.} 
\label{fig:nosal}
\end{figure*}

\begin{figure*}[!ht]
\centering
\includegraphics[width=1\textwidth, height=300px]{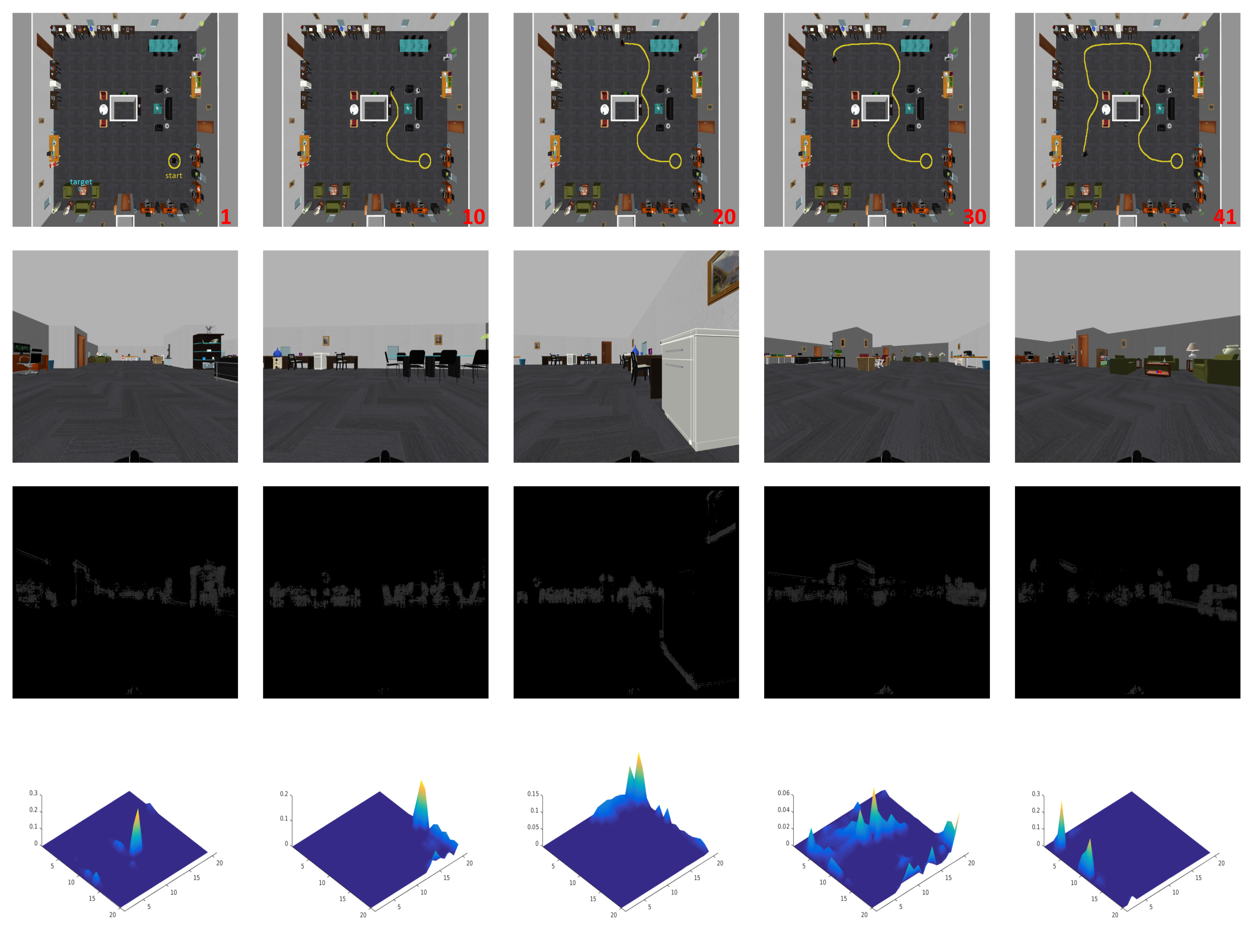} 
\caption{Experiment 1. A sample search run using the \textit{BU} method. The object of interest is the Homer toy, which is placed on the coffee table between the green sofa set at the bottom-left corner in the top-down view. From the top: top-down view, the robot's view, the bottom-up saliency map and the probability of finding the target.} 
\label{fig:aim}
\end{figure*}

\begin{figure*}[!ht]
\centering
\includegraphics[width=1\textwidth, height=300px]{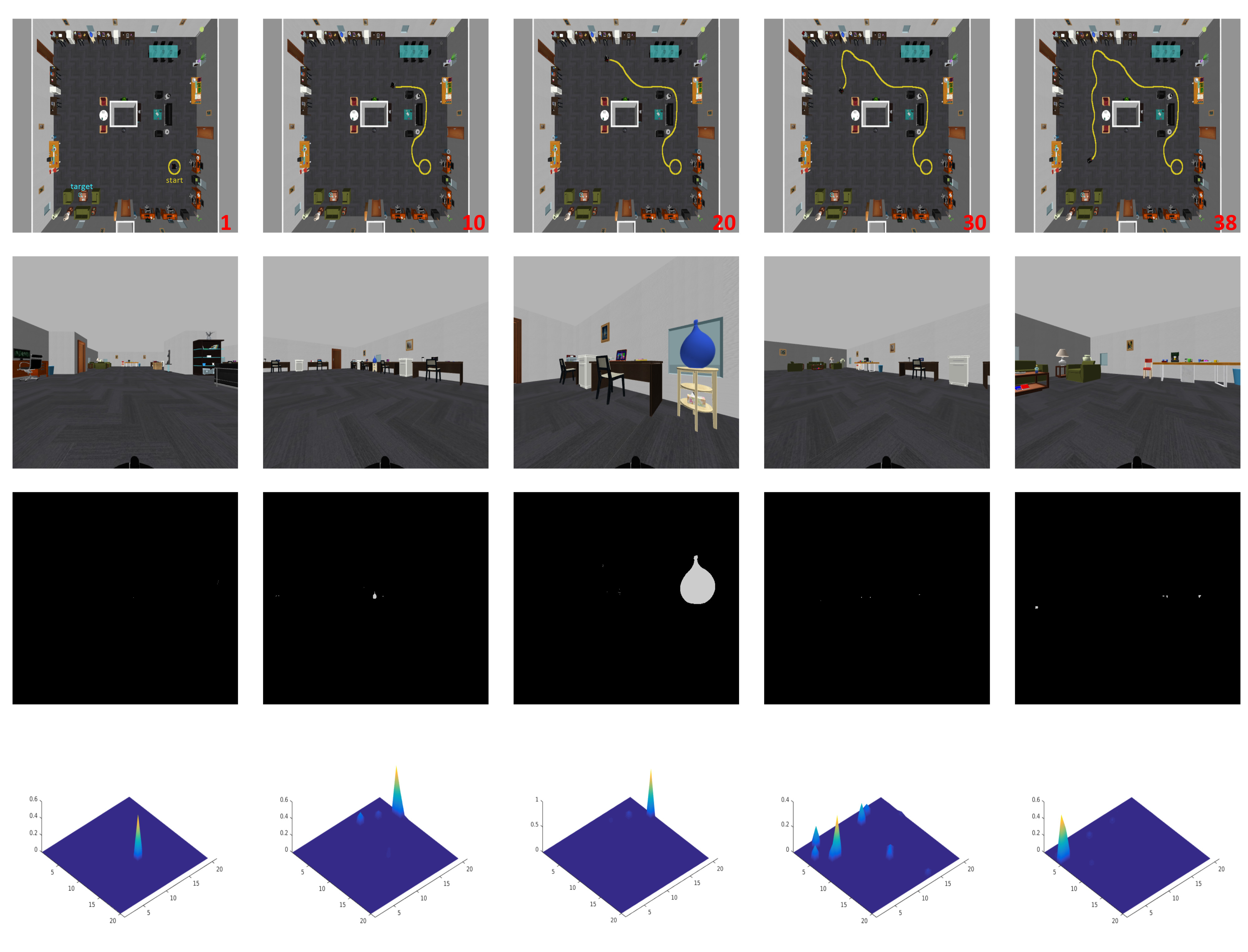} 
\caption{Experiment 1. A sample search run using the \textit{TD} method. The object of interest is the Homer toy, which is placed on the coffee table between the green sofa set at the bottom-left corner in the top-down view. From the top: top-down view, the robot's view, the bottom-up saliency map and the probability of finding the target.} 
\label{fig:bp}
\end{figure*}

\begin{figure*}[!ht]
\centering
\includegraphics[width=1\textwidth, height=300px]{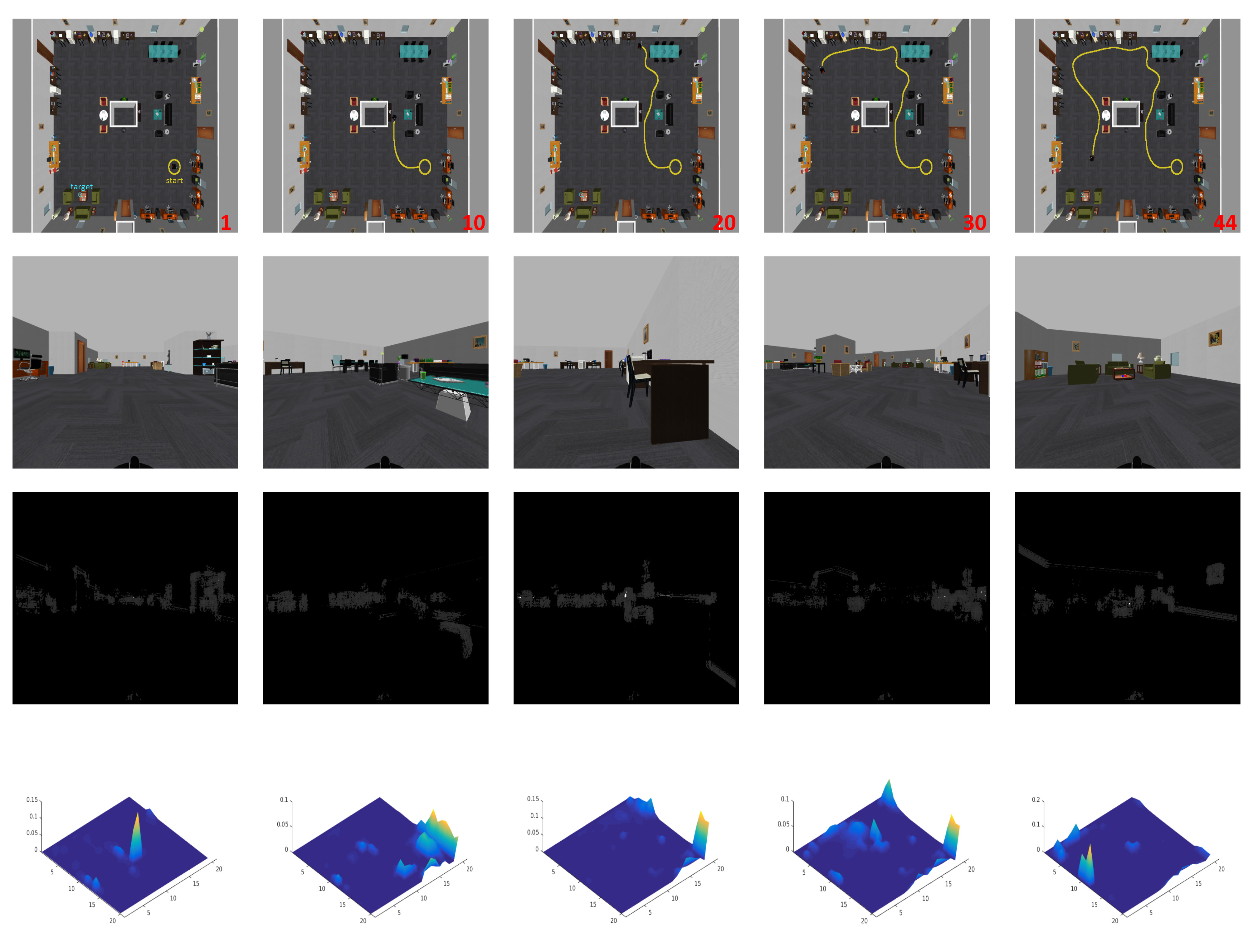} 
\caption{Experiment 1. A sample search run using the \textit{BU + TD} method. The object of interest is the Homer toy, which is placed on the coffee table between the green sofa set at the bottom-left corner in the top-down view. From the top: top-down view, the robot's view, the bottom-up saliency map and the probability of finding the target. } 
\label{fig:bu and bp}
\end{figure*}

A sample run of the proposed algorithms can be found in Figs. \ref{fig:nosal} - \ref{fig:bu and bp}. In these scenarios, the target is the Homer figurine, which is placed on the coffee table between the green sofas at the bottom-left of the environment.

\begin{table}[!hbtp]
\caption{Quantitative analysis exp. 1: \textit{NOSAL}, \textit{BU}, \textit{TD}, \textit{BU+BU$\odot$TD}, \textit{BU+TD}}
\centering
\resizebox{\columnwidth}{!}{
\small
\begin{tabular}{ccccc}
\toprule
\multicolumn{1}{c}{\textbf{Method}} & \textbf{Number of Actions} & \textbf{Search Time} & \textbf{Improvement (\%)} \\ 
\midrule
						& {\textit{mean} $\backslash$ \textit{median}}	& {\textit{mean} $\backslash$ \textit{median} (s)}	& { \textit{num acts}$\backslash$\textit{time} \% } \\ 
\midrule
{\textit{NOSAL}} &

 $114.07 \pm 86.98 \backslash 98$ & $1729.2 \pm 1148.1 \backslash 1293.6$ & - \\
\midrule

{\textbf{BU}} &
$\mathbf{66.40 \pm 61.12 \backslash 47}$ & $\mathbf{2264.8 \pm 910.8 \backslash 700.3}$ & $\mathbf{41.79\% \backslash 34.29\%}$ \\ 
\midrule

{\textit{BU+BU$\odot$TD}} &

$84.22 \pm 72.82 \backslash 57$ & $ 1659.4 \pm 1088.7 \backslash 852.1 $ &  26.17\% $\backslash$ 16.38\% \\
\midrule
{\textit{BU+TD}} &

$83.59 \pm 77.1 \backslash 55$ & $ 1031.5 \pm 1152.6 \backslash 822.3 $ &  26.72\%$\backslash$ 17.01\% \\
\midrule
{\textit{TD}} &

$79.78  \pm 78.32 \backslash 54$& $ 2856.9 \pm 1045.6 \backslash 720.9 $ &  30.07\% $\backslash$ 29.27\% \\
\bottomrule
\end{tabular}  
}
\label{table:quantitativeresults}
\end{table}
\normalsize
The result of the first experiment is summarized in Table \ref{table:quantitativeresults}. As can be seen, using any form of visual attention in visual search can significantly improve the efficiency of search. The improvement is both in terms of the number of actions performed to find the object (between  26-41\%) and the time of search (between 16-34 \%). The time improvement is lower due to the computationally expensive saliency maps of the environment and can be further improved by optimizing the implementation of the algorithms.

Despite the overall improvement, the performance of attention-based search models varies in different scenarios. In the cluttered environment, using only top-down color features to produce visual saliency is not as effective as using a combined model. The performance of the \textit{TD} algorithm changes depending on the characteristics of the target. For instance, this method performed the worst in the case of the Homer toy as shown in Figs. \ref{fig:exp1_acts} and \ref{fig:exp1_time}. This is due to the fact that this object's dominant color, blue, is similar to numerous objects in the environment such as table tops, vases, bins, shelves, etc. On the other hand, \textit{TD} achieved the best performance searching for the green glue bottle since there are fewer green objects present in the scene. 

Although, on average, the methods with bottom-up influence achieved the best results, they also performed differently when searching for different objects. For example, in the case of the Homer figurine, \textit{BU+BU$\odot$TD} had a better performance than \textit{BU+TD}. This is because in the \textit{BU+BU$\odot$TD} method,  AIM is applied to first filter the image and then the remaining parts of the image is are used to find color similarities by applying the top-down saliency model. Given that the majority of blue distractors (objects with similar colors to the target) in the scene are larger objects such as table tops, they are filtered out by the AIM algorithm because they have very low responses in the bottom-up saliency map. On the other hand, in searching for the glue bottle \textit{BU+BU$\odot$TD} does not perform well as most of the distrators are smaller objects such as beer bottles or plants. 

Averaging the results over all scenarios \textit{BU} has the best performance (of course after the search with prior knowledge). This confirms that top-down influence is limited by the environment similarities to the sought object. In a highly complex scene, using a simple color-based top-down model not only does not add any benefit, but it also may distract the attention of the robot to irrelevant objects.

\begin{figure*}[!tbp]
\centering
\subfigure[A Snapshot of the disaster environment]{\includegraphics[width=0.40\textwidth, height=160px]{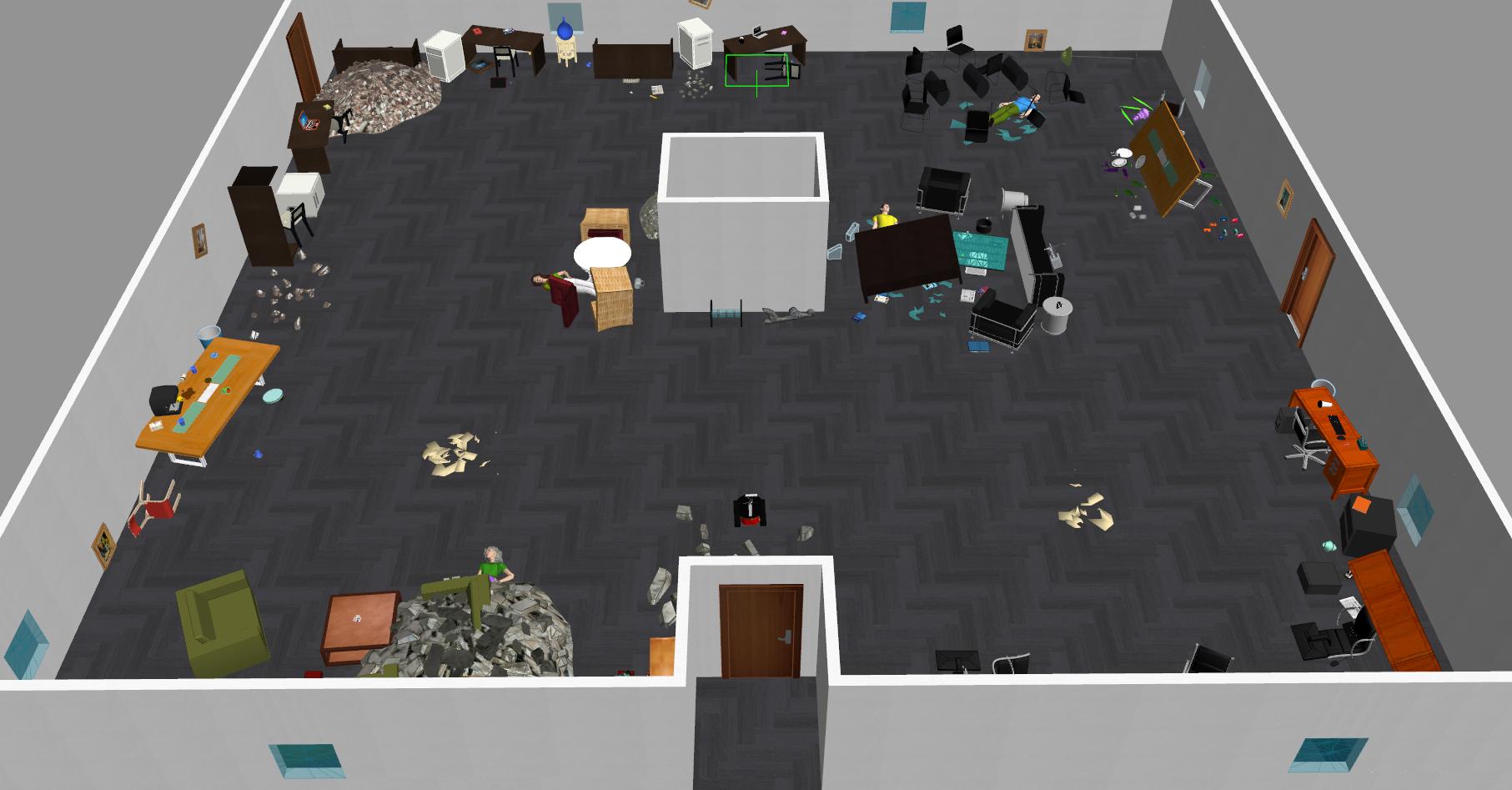} 
\label{fig:exp2_scene}}
\hspace{0.5cm}
\subfigure[Mean number of actions and time of search for each method]{\includegraphics[width=0.40\textwidth]
{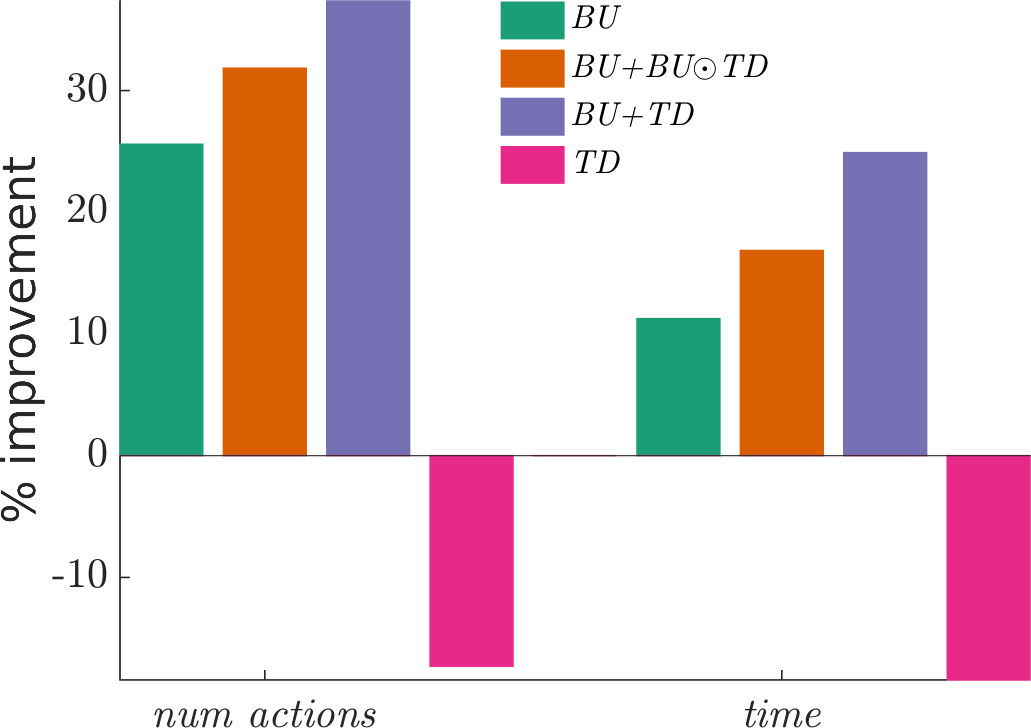}
\label{fig:exp2_result}}

\caption{Experiment 2. Search and rescue scenario. Several injured humans are in the room after an earthquake. The humans are lying on the ground and some are trapped under debris or furniture. The robot has to find all of humans as soon as possible.}
\label{fig:scenario2}
\end{figure*}

\subsection{Experiment 2. Search and Rescue in a Disaster Scene}
\label{sec:results:2}
The second experiment, depicted in Fig. \ref{fig:exp2_scene}, models a devastated building after an earthquake. The furniture is scattered around the place, the windows are broken, some objects are shattered and the environment is filled with debris and rubble. Here, the objective for the robot is to find 4 human subjects (Fig. \ref{fig:human_sample}) that are trapped under piles of furniture or debris.

We repeated the experiments by placing the robot at 7 different locations close to the doors and windows (potential entrances for search and rescue scenarios) using the proposed algorithms. A total of 35 experiments were conducted. Here we used skin color to build our top-down model. As for the optimization of the search, the number of planning steps was set to 3 and execution to 2. 

\begin{table}[!hbtp]
\caption{Quantitative analysis exp. 2: \textit{NOSAL}, \textit{BU}, \textit{TD}, \textit{BU+BU$\odot$TD}, \textit{BU+TD}}
\centering
\resizebox{\columnwidth}{!}{
\small
\begin{tabular}{ccccc}
\toprule
\multicolumn{1}{c}{\textbf{Method}} & \textbf{Number of Actions} & \textbf{Search Time} & \textbf{Improvement (\%)} \\ 
\midrule
& {\textit{mean} $\backslash$ \textit{median}}	& {\textit{mean} $\backslash$ \textit{median} (s)}	& { \textit{num acts}$\backslash$\textit{time} \% } \\ 
\midrule
{\textit{NOSAL}} &

 $98.29 \pm 48.7 \backslash 74$ & $1586.2 \pm 786.1 \backslash 1194.1.6$ & - \\
\midrule

{\textit{BU}} &
$73.14 \pm 27.34 \backslash 67$ & $1407.7 \pm 523.36 \backslash 1286.3$ & 25.58\% $\backslash$ 11.26\%  \\ 
\midrule

{\textit{BU+BU$\odot$TD}} &
$67 \pm 36.38 \backslash 78$ & $ 1319.0 \pm 658.86 \backslash 1509.1 $ &  31.83\% $\backslash$ 16.85\% \\
\midrule
{\textbf{BU+TD}} &

$\mathbf{61.57 \pm 22.27 \backslash 58}$ & $\mathbf{ 1191.4 \pm 429.57 \backslash 1119.6}$ & $\mathbf{ 37.35\% \backslash 24.89\% }$ \\
\midrule
{\textit{TD}} &

$115.28  \pm 41.88 \backslash 96$& $ 1878.5 \pm 683.0 \backslash 1563.5 $ &  -17.29\% $\backslash$ -18.42\% \\
\bottomrule
\end{tabular}  
}
\label{table:quantitativeresults_exp2}
\end{table}

As indicated in Table \ref{table:quantitativeresults_exp2}, the \textit{TD} algorithm exhibits its worst performance. This, once again, is due to high color similarity of distractors to the human skin color. In this experiment the similarity is extreme because a large number of the furniture items (Fig. \ref{fig:exp2_result}) have similar color to the subjects' skin color used for top-down saliency. The bottom-up algorithm on its own, \textit{BU}, also performed worse in comparison to the previous experiment. Here the number of distractors that can be identified as salient by the AIM algorithm is increased. For instance, debris, broken furniture or fragments of the walls on the ground can induce high saliency responses in the bottom-up map.

Overall, in this experiment the best performance was achieved by the combined models, \textit{BU+TD} and \textit{BU+BU$\odot$TD}. In these methods, the negative effects of distractors are lowered due to the fusion of saliency maps from the bottom-up and top-down models.

\section{The path to implementation}

The advantage of our simulation environment is that everything is implemented under the ROS framework. This includes all controls for navigation, localization, mapping and message passing as well as all visual processing for the saliency models. The 3D robot model (both its appearance and dynamics), the laser scanner and the camera model are also designed to be identical to the actual practical systems. Such a design approach to our simulation makes it possible to use the entire proposed system on an identical practical platform with minimum amount of modifications necessary. 

\section{Conclusion}
\label{sec:conclusion}

We have presented an active visual search approach, which improves the previous methods by grounding the decision-making on the relevant sensory information provided by the camera. It has been shown that integrating top-down and bottom-up attention within the action optimization algorithm enables the robot to react to the right stimuli in an informed manner, which produces smooth trajectories and non-greedy exploratory behaviors. In fact, our method gives the robot general responsiveness in unknown scenarios, enforces positive guidance towards stimuli that fit with the sought object, produces non-myopic behaviors and reduces the time to find the object. 

We experimentally demonstrated that by using visual saliency significant improvements, up to 40\% reduction in the number of actions performed and 35\% in overall time, can be achieved in visual search. The enhancement, however, may vary depending on the structure of the environment, type of the sought object or the type of visual saliency used. 

In a typical structured environment, bottom-up saliency information can be useful to guide the search agent to the regions of interest, e.g. tables or shelves,  that have a higher probability of containing the sought object. This is also true for the color-based top-down saliency, in particular, when the object's color is less common in the environment. However, in a more cluttered and unstructured environment these saliency models on their own may not be as effective. In some cases not only they fail to provide any informative clues, but also lead the search agent to irrelevant locations. In these scenarios, combining both bottom-up and top-down models and benefiting from the strength of both resulted in achieving a better performance.

Moreover, we showed that adding prior knowledge about the possible location of the object has outperformed dramatically all of the evaluated systems. This means that visual attention can be combined with spatial prior information to improve the efficiency of the active search.

In the future, to further validate the attention-based active search, we will implement the proposed approach on several robotic platforms to demonstrate the applicability of the methods in non-simulated scenarios. Furthermore, we will investigate other models of attention and top-down modulation to be used in the optimization function and how different perceptual embodiments affect the searching behavior.


\bibliography{references}

\end{document}